\documentclass[
]{ceurart}

\usepackage{listings}
\usepackage{inputenc}
\usepackage{booktabs}
 \usepackage{multirow}
\begin{document}

\copyrightyear{2025}
\copyrightclause{Copyright for this paper by its authors.
  Use permitted under Creative Commons License Attribution 4.0
  International (CC BY 4.0).}

\conference{MediaEval'25: Multimedia Evaluation Workshop,
  October 25--26, 2025, Dublin, Ireland and Online}

\title{LLM-based Fusion of Multi-modal Features for Commercial Memorability Prediction}

% \tnotetext[1]{You can use this document as the template for preparing your publication. We recommend using the latest version of the ceurart style.}

\author{Aleksandar Pramov}[%
    orcid=https://orcid.org/0009-0005-9049-1337,
    email=apramov3@gatech.edu,
    % url=<website>,
] 
 
% Affiliations 
\address{Georgia Institute of Technology, USA}

\begin{abstract}
This paper addresses the prediction of commercial (brand) memorability as part of \emph{``Subtask 2: Commercial/Ad Memorability"} within the \emph{``Memorability: Predicting movie and commercial memorability"} task at the \emph{MediaEval 2025} workshop competition. We propose a multimodal fusion system with a Gemma-3 LLM backbone that integrates pre-computed visual (ViT) and textual (E5) features by multi-modal projections. The model is adapted using Low-Rank Adaptation (LoRA). A heavily-tuned ensemble of gradient boosted trees serves as a baseline. A key contribution is the use of LLM-generated rationale prompts, grounded in expert-derived aspects of memorability, to guide the fusion model. The results demonstrate that the LLM-based system exhibits greater robustness and generalization performance on the final test set, compared to the baseline.\\ The paper's codebase can be found at \url{https://github.com/dsgt-arc/mediaeval-2025-memorability}
\end{abstract}

\maketitle

\section{Introduction}\label{sec:intro}
Video memorability plays a central role in various applications such as marketing and advertisement, film-making, and higher education. Modeling video memorability is particularly challenging because it requires integrating multi-channel data (visual, audio, and text) to predict a latent, unobservable, subjective characteristic of the data.

% At the same time, there is great commercial interest in understanding the drivers behind video memorability, in order to create e.g., more memorable marketing campaigns, educational videos, movies and other media \cite{martin2023video}.
% To that end, some studies suggest that certain visual elements are more likely to be memorable for later recall, despite the great deal of subjective preferences when measuring the latent memorability level of a video .
The \emph{``Memorability: Predicting movie and commercial memorability"} task at \emph{MediaEval 2025} workshop competition undertakes to study methods that can both predict memorability, as well as shed light on the drivers behind both movie and commercial memorability \cite{2025overview}. This paper deals in particular with the latter question (\emph{``Subtask 2: Commercial/Ad Memorability"}) by modeling the memorability score of commercial videos (from the financial industry) based on video and text features, as well as their brand memorability (i.e., how well a brand is remembered from a video). To that end, this year's competition dataset includes text and pre-computed video features of 424 curated YouTube commercials from financial institutions. The provided data dictates the need to integrate multi-modal features for predictive modeling: numeric (e.g., video engagement metrics), text (e.g., video titles and subtitles), and visual embeddings. One important modeling limitation was the lack of access to the raw videos, making the usage of off-the-shelf modern multimodal LLMs difficult \cite{martin2025parameter}. The approach undertaken here thus focuses on feature integration by applying both a gradient boosting model as a baseline and leveraging an LLM for the multimodal fusion of video metadata, textual embeddings, and visual features \cite{esteban2024llm}.
 
\section{Related Work}\label{sec:work}
  Previous works in video memorability prediction often relied on sophisticated feature engineering, integrating components such as deep visual features, textual semantics, and spatio-temporal information to enhance predictive capabilities $\cite{shekhar2017show, martin2024larger}$. More recently, the field has embraced leveraging the vast world knowledge and complex reasoning abilities of Multimodal Large Language Models (MM-LLMs), such as Gemma and Qwen-VL, with fine-tuning techniques such as LoRA, for tasks like human perception analysis and long-term advertisement memorability $\cite{hu2022lora, esteban2024llm, martin2025parameter,harini2025long, martin2023video}$.   

\section{Approach}\label{sec:approach}
\subsection{Data manipulation \& Evaluation setup}
The training dataset for MediaEval's 2025 Subtask 2 edition consists of a small sample of 339 videos for training and 85 videos for evaluation, sampled from the VIDEM Dataset \cite{kiziltepe2025videm} - this was the only dataset used in our work. 
% We did not use any additional external dataset for training - all the analyses were done based only on data released by the organizers. 
Moreover, multiple videos are from the same channel on very similar topics, e.g., videos on company earnings calls from different quarters within the same year. To that end, we employ two data manipulation steps on our training sample of 339 videos: First, we break down the biggest channel (``Goldman Sachs", with 23\% of all the videos) into three ``subchannels", based on a k-medoids clustering algorithm on the embedded\footnote{\url{https://huggingface.co/intfloat/e5-base-v2}} video titles. This subgrouping creates a more balanced distribution of the channels (the largest one being 37). Second, we create a nested 5-fold, grouped (by channel name) \& stratified (by 5 quantiles of the target variable) grouping of the data. 

The aim of this nested, stratified and grouped evaluation setup is to estimate the generalization error in a more robust way, while guarding against information leakage. We end up having 5 folds in the outer loop, each with 5 inner train/validation splits that will be used to elicit hyperparameters, early stopping and/or estimate the validation error.\footnote{Throughout the rest of this manuscript, "CV SRCC" refers to the averaged test error across the 5 outer folds. The performance on the 85 held-out videos from the competition is referred to as "Test SRCC."}
 
\subsection{Model features}
For the modeling, we investigated the following provided (or derived) features as input to the models, the detailed integration of which is discussed in the next section:
\begin{itemize}
\item \emph{Numerical metadata}, provided by the organizers.
\item \texttt{E5-base-v2} \emph{Embeddings} of the \emph{subtitles} (chunked and pooled), \emph{titles}, and \emph{descriptions}.
\item Pre-computed \emph{video embeddings}, provided by the organizers.
\item \emph{Subtitle summaries} of up to 1024 tokens, generated by \texttt{gemma3-4b-it-qat}, used both as \texttt{E5-base-v2} embeddings and as text for model prompts.
\item Fold-aware few-shot \texttt{gemma3-4b-it-qat} generated \emph{(brand) memorability text rationales}, based on the subtitles of the videos. The aim of the prompt was to mimic an expert system to evaluate (qualitatively) the (brand) memorability of the video, based on the subtitles only, along key characteristics such as brand integration, clarity of brand messaging, semantic richness, novelty and others.\footnote{The few-shot prompting here did not involve providing examples of full text of rationales, but there was fold-adaptive guidance with a few-shot examples of 2-3 (brand) memorability scores of the respective inner fold. These examples were chosen to be of those videos, that were the nearest neighbors by cosine similarity in the (pooled and averaged) embedding space of subtitles, titles and description.} The generated rationales were then either used as \texttt{E5-base-v2} embeddings, or in text form, to be a part of a prompt in one of the memorability predictive systems. 
\end{itemize}

\subsection{Modeling}
 We investigated two modeling architectures with (brand) memorability as a univariate target variable. The first, a baseline histogram gradient boosted tree model (\textbf{HGBT}), took as input the various text and visual embeddings, as well as the numerical metadata. This approach naturally has an overwhelming number of features for a very limited sample; thus, strong PCA reduction for each of the individual text and image feature streams was employed as a pre-processing step. The hyperparameters of the HGBT models were tuned on the inner folds using \texttt{Optuna} to optimize for Spearman correlation, and the final model for each outer fold was evaluated on the outer test split. 

The second approach delves into the complexity of joint textual and visual modeling and closely follows a previously outlined approach by Esteban et al., of using instruction-tuned LLMs as multi-modal feature integrators \cite{esteban2024llm, liu2023visual, verma2024cross}.  Our implementation, which we call \textbf{Gemma Fusion}, uses a \texttt{gemma-3-4b-it}\footnote{https://huggingface.co/google/gemma-3-4b-it} model as its backbone. We augment the model's standard text-based input with external feature streams that are projected into the LLM's embedding space as via separate trainable linear projectors, with an early fusion step at the embedding level. The textual prompt is constructed from the video's title and the aforementioned LLM-generated text, which is either a summary of the subtitles or a qualitative rationale on the video's memorability. 

In parallel, external features—including E5 embeddings for subtitles, title, and description, as well as pre-computed visual blocks through the linear projector, effectively adapting the (visual) input features to be considered as input tokens  for the LLM. The fused unified sequence is then fed to the Gemma backbone. The last hidden state from the transformer blocks is (mean or attention pooled) and passed through an MLP head producing a (brand) memorability prediction. The MLP and the projectors are trained, optimizing using an equally weighted composite loss function that is a weighted average of Mean Absolute Error (MAE) and a correlation coefficient \cite{esteban2024llm}. The gemma backbone is either frozen or Low-Rank Adaptation (LoRA) is performed on the query, key, value, and output projections in the attention layers, as well as the gate, up, and down projections of the feed-forward layers. 
   
\section{Results and Analysis}
The final results of our experiments, summarizing both the 5-fold cross-validation (CV) and the official competition test set performance, are presented in Table \ref{tab:final_publication_results}. Only the ViT embeddings were used, as analyses showed that including the other provided embeddings did not yield an added value to our model. Given the very small size of the sample, we decided to pick a visual embedding model, the performance of which on such task is grounded in previous literature \cite{esteban2024llm}.  

One key highlight is that, despite the extensive efforts, overfitting did occur quite substantially for the HGBT models. The performance on the competition dataset collapsed, compared to the CV procedure. This suggests that while the feature engineering was effective on the development data, the model ultimately overfit to its specific characteristics.
% This leaves room for improvement, potentially with ensemble boosted tree algorithms which allow for more regularization, e.g. \texttt{xgboost}.

In contrast, the Gemma Fusion model exhibited greater robustness and superior generalization - the LLM-based feature integration approach exhibited much better performance in three of the four submissions. Using LoRA proved beneficial in multiple ablations (not all shown here) and thus all final Gemma employed LoRA with a rank of 32, an alpha of 32, and a dropout of 0.15, applied to both attention projections and feed-forward layers. The best-performing setups consistently used mean pooling and included the ViT visual block and all three E5 text streams (subtitles, title, and description). Notably, the performance of the prompt content was target-dependent. On the final test set, Brand Memorability benefited more from rationales as part of the prompt (0.122 vs. 0.112 SRCC), while Memorability Score performed substantially better using the subtitle summary (0.131 vs. 0.018 SRCC).

\begin{table*}[t]
\centering
\caption{\scriptsize{\textbf{Cascaded ablation and final submission results}\newline Comparing the mean of \emph{outer fold} 5-fold cross-validation (CV) scores with official competition test set scores. Final submitted models are indicated by a result in the Test SRCC column. \textbf{SumEmb}=E5(Summaries), \textbf{RatEmb}=E5(Rationales) embeddings. \textbf{Attn./Mean}=Pooling. \textbf{P}=Prompt content (Rat.=Rationales, Sum.=Summaries). See main text for detailed information the models and the nested CV construction.}}
\label{tab:final_publication_results}
\footnotesize  
\setlength{\tabcolsep}{4pt}  
\begin{tabular}{lllccc}
\toprule
\textbf{Target} & \textbf{Model} & \textbf{Config.} & \textbf{CV SRCC } & \textbf{CV RMSE} & \textbf{Test SRCC} \\
\midrule

% Brand Memorability Section
\multirow{8}{*}{\begin{tabular}[c]{@{}l@{}}Brand \\ Memorability\end{tabular}} & \multirow{4}{*}{HGBT} & Base (E5(Text)+Numeric) & 0.0021 & 0.1662 & - \\
& & Ablation (Base + ViT) & 0.0304 & 0.1569 & - \\
& & \textbf{+ SumEmb} & \textbf{0.1181} & \textbf{0.1547} & \textbf{0.019} \\
& & + RatEmb & 0.0504 & 0.1574 & - \\
\cmidrule(l){2-6}
& \multirow{4}{*}{Gemma Fusion} & Base (P: Rat) & 0.0588 & 0.1701 & - \\
& & Ablation (+E5(Text)+ViT, Attn.) & 0.0223 & 0.1534 & - \\
& & Final (E5(Text)+ViT, Mean, P: Rat.) & 0.1569 & 0.1506 & \textbf{0.122} \\
& & Final (E5(Text)+ViT, Mean, P: Sum.) & 0.1627 & 0.1515 & \textbf{0.112} \\
\midrule

% Memorability Score Section
\multirow{6}{*}{\begin{tabular}[c]{@{}l@{}}Memorability \\ Score\end{tabular}} & \multirow{4}{*}{HGBT} & Base (E5(Text)+Numeric) & 0.1190 & 0.1490 & - \\
& & Ablation (Base + ViT) & 0.1577 & 0.1493 & - \\
& & + SumEmb & 0.1955 & 0.1436 & - \\
& & \textbf{+ RatEmb} & \textbf{0.2402} & \textbf{0.1425} & \textbf{0.089} \\
\cmidrule(l){2-6}
& \multirow{2}{*}{Gemma Fusion} & Final (E5(Text)+ViT, Mean, P: Rat.) & 0.1165 & 0.1580 & \textbf{0.018} \\
& & Final (E5(Text)+ViT, Mean, P: Sum.) & 0.0849 & 0.2158 & \textbf{0.131} \\
\bottomrule
\end{tabular}
\end{table*}

\section{Discussion and Outlook}
The small sample size proved to be a significant challenge for the task. On the one hand, a simpler model like HGBT still overfit and did not prove to be a good baseline on the competition dataset. On the other hand, our LLM-multimodal fusion approach exhibited much more robust performance but still indicates issues with the training stability, as evidenced by the collapse of the rationales run for the memorability score (from 0.1165 on the CV SRCC to 0.018 on the test). Nonetheless, the LLM-multimodal fusion approach managed to lift the the baseline performance both in absolute terms and in terms of stability of the system. In addition, the generated LLM rationales introduced a novel and alternative way to prompt such a fused model, which did lift the performance for the brand memorability.

Future work will involve the addition of additional datasets for training, e.g. \texttt{memento10k} which should improve the overall stability, and possibly performance of the model. In addition, crafting better, (brand) memorability-expert oriented prompts for the model could be another promising avenue for further research.  Lastly, throughout the whole analysis we did not leverage specific domain information - models which are fine-tuned on textual data from financial domain could be a better choice and have the potential to bring added value.

  \begin{acknowledgments}
We thank the DS@GT team for providing valuable comments and suggestions.
This research was supported in part through research cyberinfrastructure resources and services provided by the Partnership for an Advanced Computing Environment (PACE) at the Georgia Institute of Technology, Atlanta, Georgia, USA.
\end{acknowledgments}

\section*{Declaration on Generative AI}

 During the preparation of this work, the author(s) used Gemini 2.5 in order to perform grammar and spelling check. After using these tool(s)/service(s), the author(s) reviewed and edited the content as needed and take(s) full responsibility for the publication’s content. 

\def\bibfont{\small}  

\bibliography{references} 

\end{document}